\pdfoutput=1
\RequirePackage[T1]{fontenc}
\documentclass[journal]{IEEEtran}
\ifdefined\pdfinfoomitdate\pdfinfoomitdate=1\fi
\ifdefined\pdfsuppressptexinfo\pdfsuppressptexinfo=-1\fi
\ifdefined\pdftrailerid\pdftrailerid{}\fi
\usepackage{cite}
\usepackage{amsmath,amssymb}
\usepackage{graphicx}
\usepackage{booktabs,array}
\usepackage{algorithm,algpseudocode}
\usepackage{url}
\usepackage[hidelinks]{hyperref}
\hypersetup{
  pdftitle={GARDEN: Generalist Autonomous Reward Design Through Evidence Navigation for Cooperative Multi-Agent Reinforcement Learning},
  pdfsubject={Autonomous reward engineering and within-family reward transfer},
  pdfkeywords={Cooperative systems, large language models, multi-agent systems, reinforcement learning, transfer learning}
}
\hypersetup{pdfauthor={Dogan Urgun}}
\graphicspath{{./}{figures/}}
\newtheorem{definition}{Definition}
\newtheorem{problem}{Problem}
\newcommand{\ms}[2]{$#1\pm#2$}
\newcommand{\bms}[2]{{\boldmath$#1\pm#2$}}
\newcommand{\policyseeds}{10}
\newcommand{\spreadnote}{Performance values are means $\pm$ standard errors over \policyseeds{} matched policy seeds.}

\begin{document}
\title{GARDEN: Generalist Autonomous Reward Design Through Evidence Navigation for Cooperative Multi-Agent Reinforcement Learning}
\author{Dogan~Urgun
\thanks{D. Urgun is with the Department of Electrical and Electronics Engineering, Karabuk University, 78050 Karabuk, T\"{u}rkiye (e-mail: \mbox{durgun@karabuk.edu.tr}).}}
\markboth{Preprint}{Urgun: GARDEN for Cooperative Multi-Agent Reinforcement Learning}
\maketitle

\begin{abstract}
Cooperative multi-agent reinforcement learning (MARL) enables teams of agents, such as robots and autonomous vehicles, to learn coordinated behavior, but what such teams learn is largely determined by their reward function. Because task success alone provides too little guidance for learning, intermediate rewards are usually designed by hand, and this effort must often be repeated whenever the environment or the team changes. Large language models (LLMs) can automate the effort by writing reward functions as code, yet the resulting rewards are typically designed and tested in a single setting. To enable their reuse across settings, GARDEN (Generalist Autonomous Reward Design through Evidence Navigation) is proposed, in which one reward program is developed for a family of related scenarios and transferred, unchanged, to scenarios that were not available during its design. During development, environment code, training records, and agent behavior are investigated by an LLM reward engineer, and draft rewards are tested in trial training runs, while the learning algorithm and the rule for selecting the final reward are held fixed. On unseen scenarios of the Overcooked and SMAX benchmarks, task returns were increased by 58.2\% and 52.5\% over the default rewards on two kitchen layouts, and the mean win rate in two combat scenarios was raised from 51.0\% to 89.9\%. Under matched candidate and probe budgets, this exceeded the 84.2\% and 79.2\% obtained with fixed diagnostic feedback and Eureka-inspired reward reflection. These results show that a single autonomously engineered reward can be reused effectively across related cooperative scenarios.
\end{abstract}
\begin{IEEEkeywords}
Cooperative systems, large language models, multi-agent systems, reinforcement learning, transfer learning.
\end{IEEEkeywords}

\section{Introduction}
\label{sec:intro}
\IEEEPARstart{M}{ulti-agent} reinforcement learning (MARL) is used to train teams of agents that cooperate toward a shared objective \cite{gronauer2022}, with applications such as multi-robot coordination \cite{zhu2025} and vehicle platooning \cite{wei2026pcrd}. In such systems, the reward function specifies what the team should achieve and, implicitly, how credit for a joint outcome is assigned to individual agents and decisions \cite{wong2023}. When only task completion is rewarded, this assignment is difficult, because a successful outcome may result from a long sequence of interdependent actions by several agents.

This difficulty is commonly addressed by reward shaping, in which intermediate learning signals are added to the task reward \cite{ng1999}. Effective shaping, however, is obtained only through domain expertise and repeated trial and error, and unintended behavior may be induced in a new setting by a reward that was effective in another \cite{hadfield2017}. For cooperative systems that must operate under varying environments and team compositions, reward design therefore becomes a recurring engineering cost.

Large language models (LLMs) have recently been used to automate reward design by writing reward functions as executable code. Reward code has been generated from environment abstractions and refined with human feedback \cite{xie2024}, improved automatically through reflection on training statistics \cite{ma2024}, and optimized through agentic search over evaluation records \cite{li2026arbo}. Two limitations of current approaches motivate this work. First, a generated reward is typically designed and evaluated within a single task or scenario. Whether it remains useful when conditions change, for example under a different spatial layout or team composition, is rarely examined, so each new configuration may require a new search. Second, the evidence used to refine a reward is usually defined in advance by the method designer, typically as summary statistics of training. Coordination failures that are not reflected in these statistics remain undiagnosed, because the underlying behavior and learned policies are not examined.

\begin{figure*}[t]
\centering
\includegraphics[width=\textwidth]{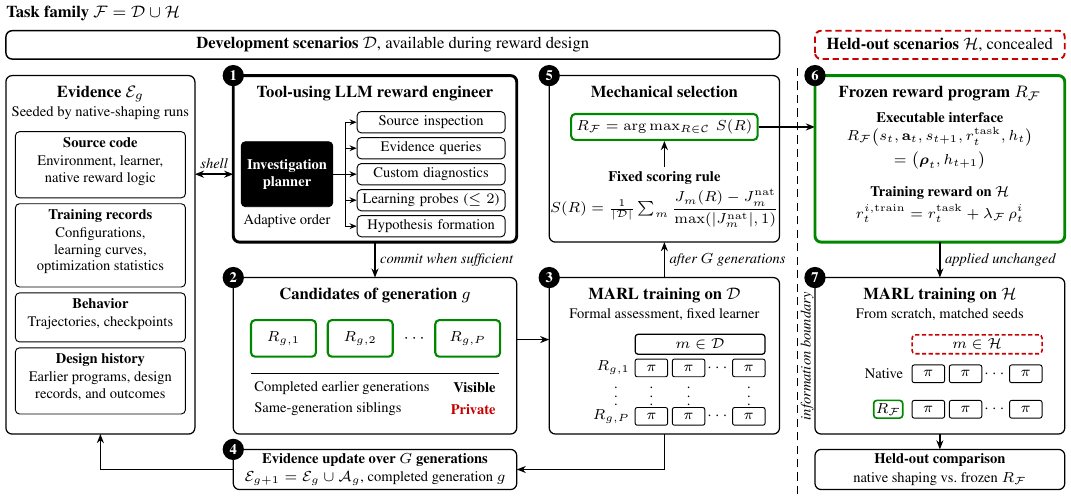}
\caption{Overview of GARDEN for one task family. Candidates are developed in independent sessions from shared development evidence and private learning probes (1, 2), assessed with the fixed learner (3), and added to the evidence (4). After $G$ generations, one program is selected by a fixed rule (5), frozen (6), and used to train new policies on the held-out scenarios, with native shaping as the reference (7). Green marks reward programs, and red marks information withheld from the reward engineer.}
\label{fig:overview}
\end{figure*}

To address these limitations, GARDEN (Generalist Autonomous Reward Design through Evidence Navigation) is proposed. An overview of the method is given in Fig.~\ref{fig:overview}. One executable reward program is developed for a family of related scenarios from several development scenarios. The choice of evidence, the construction of analyses, and the testing of draft rewards through learning probes are delegated to a tool-using LLM reward engineer, whereas the learning algorithm and the selection rule are fixed. The selected program is then frozen and used, without modification, to train new policies on held-out scenarios that were concealed during its design.

The main contributions of this paper are as follows.
\begin{enumerate}
\item Reward design is formulated at the level of a cooperative task family. A development-to-transfer protocol is introduced, in which a single executable reward program is developed on several development scenarios, frozen, and evaluated by training new policies on scenarios concealed during its design. The generalization of the reward is thereby measured separately from that of a trained policy.
\item Autonomous evidence navigation is introduced for LLM-based reward engineering. Instead of a fixed feedback report, the source code, training artifacts, and behavioral evidence to be examined are selected by the LLM reward engineer, analyses are constructed autonomously, and design hypotheses are tested through learning probes, while the learner and a mechanical selection rule are held fixed.
\end{enumerate}

To support reproducibility, the implementation, prompts, and experimental configuration are available at \url{https://github.com/DoganUrgun/GARDEN}. The remainder of this paper is organized as follows. Related work is reviewed in Section~\ref{sec:related}. The proposed method is presented in Section~\ref{sec:method}, and the experimental design is described in Section~\ref{sec:impl}. The results are reported and discussed in Section~\ref{sec:results}, and conclusions are drawn in Section~\ref{sec:conclusion}.

\section{Related Work}
\label{sec:related}
Work related to GARDEN spans reward shaping, LLM-based reward engineering, and learning signals for cooperative MARL. Across these lines of work, a reward is generally designed for and assessed on its target task, and the evidence available for its design is fixed by the method. The evidence used by representative LLM-based approaches is compared in Table~\ref{tab:positioning}.

\begin{table*}[t]
\centering
\caption{Evidence Available to the LLM in Representative Reward-Design Approaches}
\label{tab:positioning}
\footnotesize
\setlength{\tabcolsep}{8pt}
\renewcommand{\arraystretch}{1.12}
\begin{tabular}{@{}>{\raggedright\arraybackslash}p{4.0cm}>{\raggedright\arraybackslash}p{1.9cm}>{\raggedright\arraybackslash}p{10.2cm}@{}}
\toprule
Approach & Agents & Main source of reward-design evidence\\
\midrule
Text2Reward \cite{xie2024} & Single & Environment abstractions and human feedback\\
Eureka \cite{ma2024} & Single & Environment code and reward-component statistics from training\\
Self-alignment \cite{zeng2024} & Single & Comparisons of behavior under candidate rewards\\
RF-Agent \cite{gao2025} & Single & Earlier designs and training feedback organized by tree search\\
Reward Design Agent \cite{lee2026} & Single & Trajectory assessment by a vision-language model\\
ARBO \cite{li2026arbo} & Single, multi & Source inspection and tool-based access to optimization history\\
PCRD \cite{wei2026pcrd} & Multi & Fixed code-analysis questions with reward and objective curves\\
\textbf{GARDEN} & \textbf{Multi} & \textbf{Source code, raw learner artifacts, custom analyses, and private learning probes across development scenarios}\\
\bottomrule
\end{tabular}
\end{table*}

\subsection{Reward Shaping and Transfer}
Potential-based shaping preserves optimal policies under specific conditions \cite{ng1999}. Rewards can also be inferred from expert demonstrations \cite{abbeel2004} or learned from human preferences \cite{christiano2017}. Shaping knowledge has been transferred across tasks by meta-learning a potential function over a task distribution \cite{zou2021}. In cooperative MARL, credit assignment has instead been built into the learner through counterfactual baselines \cite{foerster2018} or value factorization \cite{rashid2018}. These approaches depend on demonstrations, human judgments, a predefined reward form, or changes to the learning algorithm. In GARDEN, the learner is left unchanged, and an executable reward program is synthesized from source code and empirical evidence and reused across scenarios with different geometry or team structure.

\subsection{LLM-Based Reward Engineering}
Text2Reward generated dense reward code from environment abstractions and refined it with human feedback \cite{xie2024}. Eureka evolved reward code through reflection on reward-component statistics collected during training \cite{ma2024}. The search was subsequently improved through reward-structure evolution and parameter alignment \cite{li2025rstar}, language-agent tree search \cite{gao2025}, self-alignment from execution results \cite{zeng2024}, and trajectory evaluation by a vision-language model \cite{lee2026}. In these methods, the form of the feedback is fixed by the method design, and each reward is evaluated on the task for which it was designed. In ARBO, the closest approach, reward discovery is organized as an agentic black-box optimization loop with source inspection, tools, and persistent evaluation records \cite{li2026arbo}. The choice of evidence is thus already delegated to an agent, but the records are limited to candidate scores, per-term training curves, and error tracebacks, and each reward is evaluated within the task for which it was searched. In GARDEN, raw trajectories and checkpoints are additionally examined, draft rewards are tested through private learning probes, one program is selected jointly across development scenarios, and its unchanged source is evaluated on scenarios withheld from the search.

\subsection{Learning Signals for Cooperative MARL}
In cooperative MARL, LLMs have been used to derive latent rewards for credit assignment \cite{qu2025}, to generate rewards and prior policies for multi-robot cooperation \cite{zhu2025}, to provide high-level guidance to hierarchical learners \cite{bai2026}, to design incentive-aware rewards \cite{urgun2026}, to optimize learning signals adaptively \cite{wu2026}, and to design rewards for cooperative platoon coordination \cite{wei2026pcrd}. The incentive-aware study in \cite{urgun2026} also synthesizes executable reward programs for cooperative MARL and selects them by sparse return, but a fixed set of incentive diagnostics is returned to the LLM and each reward is evaluated in the scenario used for its design, whereas in GARDEN the diagnostic process is self-directed and a single reward is transferred unchanged across a task family. These methods often modify additional learning components, such as reward redistribution or policy priors, and their signals are typically developed and evaluated within the scenarios in which they are used. MARL benchmarks, by contrast, increasingly emphasize generalization to unseen configurations, for example through procedurally generated combat scenarios \cite{ellis2023} and unseen kitchen layouts and partners \cite{ruhdorfer2025}. In GARDEN, only the reward is varied, the reward is frozen before transfer, and its value is measured on held-out scenarios.

\section{Methodology}
\label{sec:method}
\subsection{Task-Family Reward Design}
Each cooperative scenario $m$ is modeled as a decentralized partially observable Markov decision process \cite{oliehoek2016} with $N_m$ agents and episode limit $T_m$. At time $t$, action $a^i_t$ is selected by agent $i$ from observation $o^i_t$, the joint action $\mathbf{a}_t$ advances the state from $s_t$ to $s_{t+1}$, and a shared sparse task reward $r^{\mathrm{task}}_t$ is received. Performance is measured by
\begin{equation}
J_m(\pi)=\mathbb{E}_{\pi}\left[\sum_{t=0}^{T_m-1}r^{\mathrm{task}}_t\right],
\label{eq:task_return}
\end{equation}
irrespective of the shaping used during training.

\begin{definition}[Cooperative task family]
A task family $\mathcal{F}$ is a set of scenarios with shared task semantics and interaction mechanics but different structural properties, such as layout, team size, or unit composition. It is partitioned into a development set $\mathcal{D}$ and a held-out set $\mathcal{H}$.
\end{definition}

\begin{definition}[Reward program]
A reward program is a deterministic executable function
\begin{equation}
(\boldsymbol{\rho}_t,h_{t+1})=R(s_t,\mathbf{a}_t,s_{t+1},r^{\mathrm{task}}_t,h_t),
\label{eq:reward_program}
\end{equation}
where $\boldsymbol{\rho}_t=[\rho^1_t,\ldots,\rho^{N_m}_t]$ contains per-agent shaping values and $h_t$ is a bounded memory reset at episode boundaries.
\end{definition}

No scenario identifier is supplied, although scenario identity may be inferred from geometry or team composition, so generalization is assessed empirically. Two training rewards are compared,
\begin{equation}
r^{i,\mathrm{train}}_t=
\begin{cases}
r^{\mathrm{task}}_t+r^{i,\mathrm{nat}}_t, & \text{native shaping},\\
r^{\mathrm{task}}_t+\lambda_{\mathcal{F}}\rho^i_t, & \text{generated reward},
\end{cases}
\label{eq:training_rewards}
\end{equation}
where $r^{i,\mathrm{nat}}_t$ is the built-in shaping of the environment and $\lambda_{\mathcal{F}}>0$ is a fixed family weight. The native shaping may be inspected as source code but is never added to, imported by, or called from a generated program.

\begin{problem}[Task-family reward design]
Given development evidence only, a program $R$ is sought that maximizes
\begin{equation}
\frac{1}{|\mathcal{H}|}\sum_{m\in\mathcal{H}}J_m(\pi^R_m),
\label{eq:objective}
\end{equation}
where $\pi^R_m$ is trained from random initialization on scenario $m$ with the generated reward in \eqref{eq:training_rewards}.
\end{problem}
Because held-out outcomes are unavailable during design, the objective is approached through development search and a predetermined selection rule, and no transfer guarantee is implied.

\subsection{Autonomous Evidence Navigation}
\label{subsec:navigation}
The objective, information boundary, reward interface, and learner are fixed by the protocol, and the investigation within this boundary is delegated to a tool-using LLM. Environment and learner source, native reward logic, training configurations, trajectories, losses, learning curves, checkpoints, and completed design records are accessible through a constrained shell, in which files can be inspected, commands executed, and custom analysis programs constructed. No diagnostic checklist or fixed set of statistics is prescribed. A bounded number of learning probes can also be requested, in which a draft reward is trained on a chosen development scenario with the learner, budget, and seed of the formal assessment. Each committed program is accompanied by a short design report of the evidence examined, the proposed mechanism, and its limitations.

\subsection{Independent Candidates and Shared Evidence}
\label{subsec:search}
Programs are developed over $G$ generations of $P$ candidates. With $\mathcal{E}_g$ the completed evidence at the start of generation $g$ and $\mathcal{Q}_{g,p}$ the private evidence acquired by candidate $p$, including its probes, the information set of the candidate is
\begin{equation}
\mathcal{I}_{g,p}=\mathcal{E}_g\cup\mathcal{Q}_{g,p}.
\label{eq:info_sets}
\end{equation}
Each candidate is developed in an independent session, so the programs, reports, probes, and assessments of its siblings are withheld, whereas completed earlier generations are visible. After commitment, each candidate is assessed on every development scenario, and a private probe is reused only when it exactly matches the required assessment. The evidence is then updated as
\begin{equation}
\mathcal{E}_{g+1}=\mathcal{E}_g\cup\mathcal{A}_g,
\label{eq:evidence_update}
\end{equation}
where $\mathcal{A}_g$ contains the completed designs and results of generation $g$. The complete procedure is illustrated in Fig.~\ref{fig:overview} and summarized in Algorithm~\ref{alg:garden}.

\begin{algorithm}[t]
\caption{GARDEN reward engineering and transfer}
\label{alg:garden}
\begin{algorithmic}[1]
\Require Development set $\mathcal{D}$, held-out set $\mathcal{H}$, fixed learner, $G$ generations, $P$ candidates
\State Native policies are trained on $\mathcal{D}$, and $\mathcal{E}_1$ is recorded
\For{$g=1,\ldots,G$}
  \For{$p=1,\ldots,P$, in independent sessions}
    \State Evidence is investigated, and private probes may be run
    \State $R_{g,p}$ and its report are committed and validated
  \EndFor
  \State Each $R_{g,p}$ is assessed on every $m\in\mathcal{D}$
  \State The evidence is updated to $\mathcal{E}_{g+1}$ by \eqref{eq:evidence_update}
\EndFor
\State $R_{\mathcal{F}}$ is selected by \eqref{eq:selection_score} and \eqref{eq:selected_reward}, frozen, and hashed
\For{each $m\in\mathcal{H}$ and seed}
  \State Native and $R_{\mathcal{F}}$ policies are trained from scratch
\EndFor
\end{algorithmic}
\end{algorithm}

\subsection{Mechanical Selection and Transfer}
\label{subsec:selection}
After all $PG$ candidates in $\mathcal{C}$ have been assessed, one program is selected by
\begin{equation}
S(R)=\frac{1}{|\mathcal{D}|}\sum_{m\in\mathcal{D}}
\frac{J_m(R)-J^{\mathrm{nat}}_m}{\max(|J^{\mathrm{nat}}_m|,1)},
\label{eq:selection_score}
\end{equation}
\begin{equation}
R_{\mathcal{F}}=\arg\max_{R\in\mathcal{C}}S(R),
\label{eq:selected_reward}
\end{equation}
where $J_m(R)$ is the mean development sparse return under $R$ and $J^{\mathrm{nat}}_m$ is the native reference. Relative improvement is used when $|J^{\mathrm{nat}}_m|\ge1$, and absolute improvement otherwise, as for win rates. Ties are resolved by the worst-scenario normalized improvement, then the mean sparse return, then a deterministic identifier, so that neither LLM preference nor held-out results enter the selection.

The selected source is frozen and hashed, and native and generated-reward policies are then trained from scratch on the held-out scenarios under identical settings. Because no policy parameters are transferred, the measured generalization is that of a learning signal. Interface compliance, admissible dependencies, and finite execution are verified before any training, and the information boundary is protected by sanitized workspaces, restricted commands, and logged actions.

\section{Experimental Design}
\label{sec:impl}
\subsection{Environments and Rewards}
\label{subsec:envs}
All experiments were conducted in JaxMARL, a library of multi-agent environments and algorithms implemented in JAX for fast, parallel training \cite{rutherford2024}. Two task families of increasing complexity were used. Overcooked \cite{carroll2019} is a two-agent cooking task, in which ingredients must be collected, cooked, and delivered, and in which the kitchen layout determines how the two agents must coordinate. Each agent observes the full kitchen. SMAX \cite{rutherford2024} is a JAX reimplementation of the StarCraft Multi-Agent Challenge \cite{samvelyan2019}, a widely used cooperative combat benchmark, in which a team of allied units must defeat enemy units controlled by built-in heuristics. Its scenarios differ in team size, unit types, and numerical balance, and each unit observes only nearby units.

The development and held-out scenarios of both families are listed in Table~\ref{tab:scenarios}, and a separate reward program was developed for each family. The four held-out scenarios are shown in Fig.~\ref{fig:heldout}. They differ from the development scenarios in spatial layout, in team size, or in the composition of the opposing teams.

\begin{table*}[t]
\centering
\caption{Development and Held-Out Scenarios}
\label{tab:scenarios}
\footnotesize
\setlength{\tabcolsep}{7pt}
\renewcommand{\arraystretch}{1.1}
\begin{tabular}{@{}lll@{\hspace{30pt}}lll@{}}
\toprule
\multicolumn{3}{@{}c}{Overcooked} & \multicolumn{3}{c@{}}{SMAX}\\
\cmidrule(r{18pt}){1-3}\cmidrule(l){4-6}
Scenario & Role & Agents & Scenario & Role & Allied vs.\ enemy units\\
\midrule
Cramped Room & Development & 2 & 3s5z vs 3s6z & Development & 8 vs.\ 9\\
Bottleneck (adapted layout) & Development & 2 & SMACv2 5 units & Development & 5 vs.\ 5\\
Coordination Ring & Development & 2 & 5m vs 6m & Development & 5 vs.\ 6\\
Forced Coordination & Held-out & 2 & 10m vs 11m & Held-out & 10 vs.\ 11\\
Counter Circuit & Held-out & 2 & SMACv2 10 units & Held-out & 10 vs.\ 10\\
\bottomrule
\end{tabular}
\par\vspace{3pt}
\parbox{\textwidth}{\footnotesize Only allied SMAX units are trained. Unit types and start positions are randomized in every episode of the SMACv2 scenarios.}
\end{table*}

\begin{figure*}[t]
\centering
\setlength{\tabcolsep}{4pt}
\begin{tabular}{cccc}
\includegraphics[height=2.75cm]{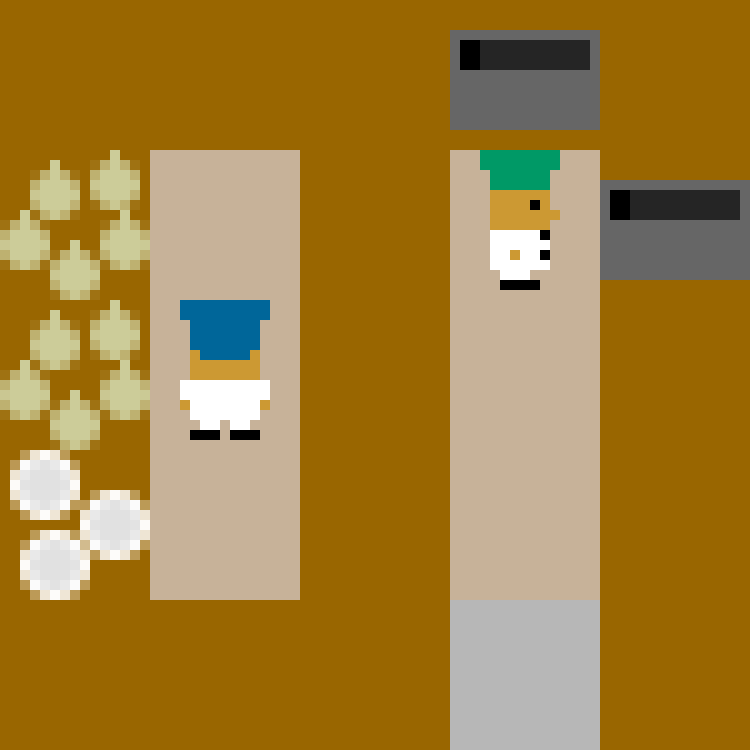} &
\includegraphics[height=2.75cm]{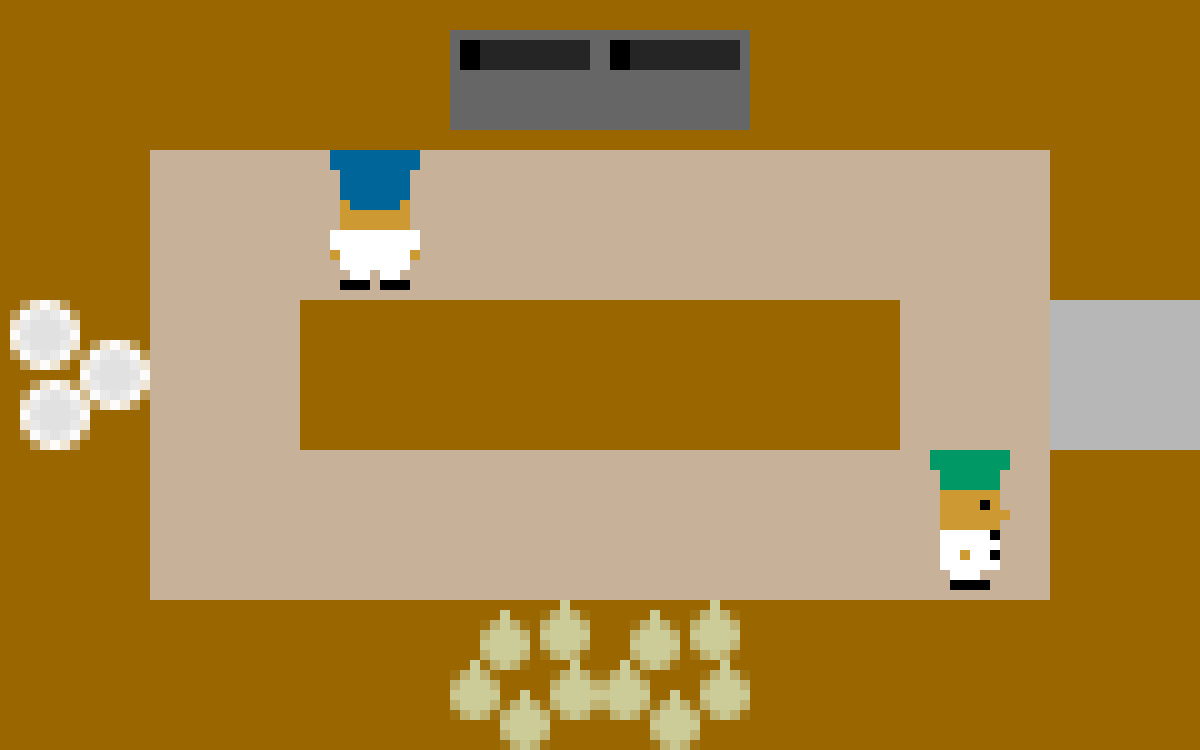} &
\includegraphics[height=2.75cm]{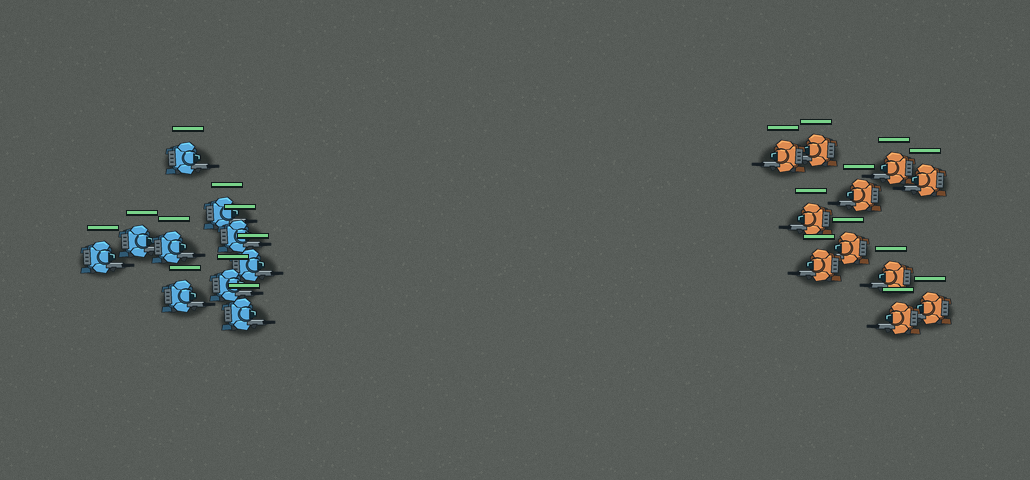} &
\includegraphics[height=2.75cm]{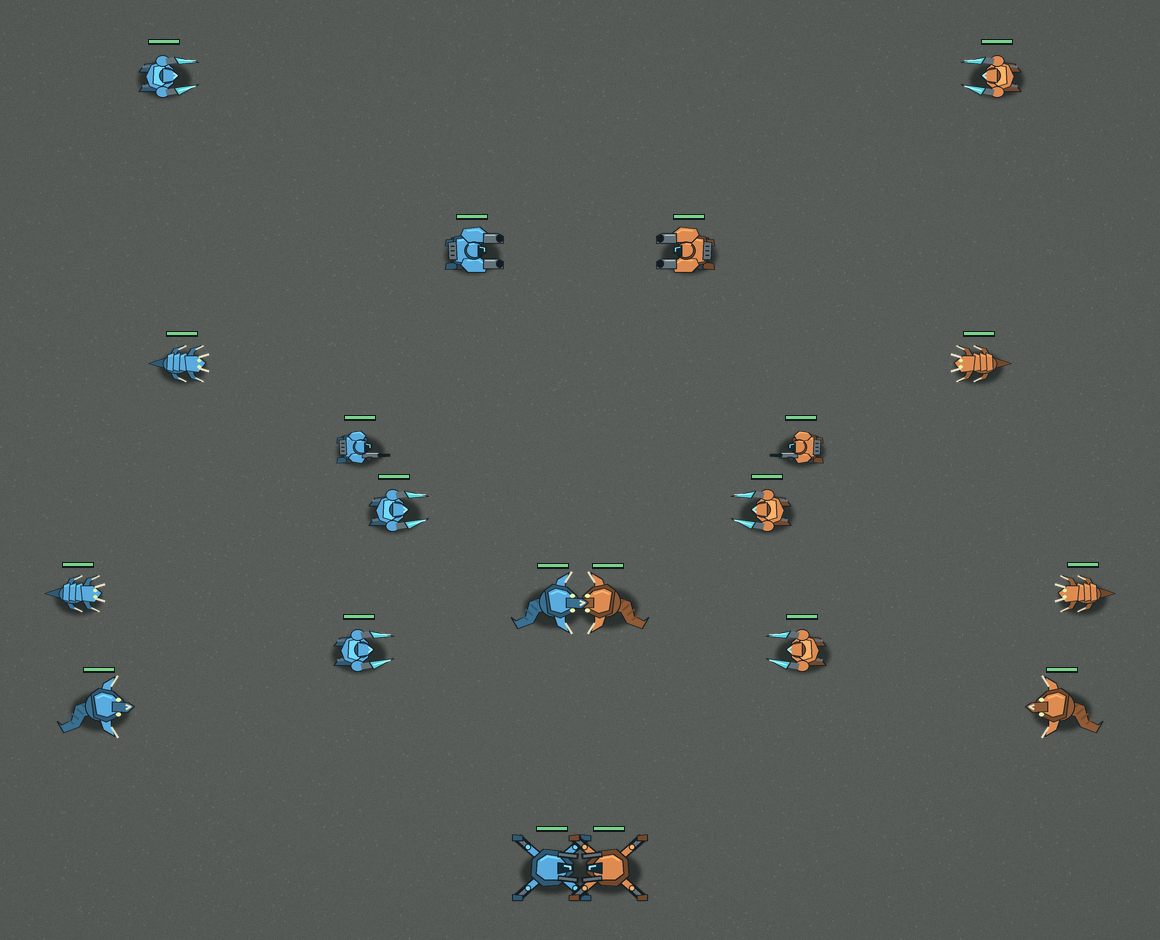}\\[1pt]
\footnotesize (a) Forced Coordination & \footnotesize (b) Counter Circuit & \footnotesize (c) 10m vs 11m & \footnotesize (d) SMACv2 10 units
\end{tabular}
\caption{Held-out scenarios. (a), (b) Overcooked layouts, in which two cooks must coordinate across a dividing counter or around a central counter. (c), (d) SMAX battles with allied units in blue and enemy units in orange, against a larger marine team in (c) and with randomized unit types and positions in (d).}
\label{fig:heldout}
\end{figure*}

The native reward of each environment is its default reward, which combines a sparse task reward with built-in shaping. In Overcooked, the task reward grants 20 points to the team for each delivered dish, and the built-in shaping rewards the acting agent for intermediate steps: 3 for placing an ingredient into a pot, 3 for picking up a dish when one is needed, and 5 for picking up a cooked soup. In SMAX, the task reward is 1 when all enemies are eliminated while at least one ally survives, so its mean over episodes equals the win rate, and the built-in shaping rewards reductions in enemy health. Performance was always measured with the task reward alone. In the generated condition, the built-in shaping was replaced by the generated program, weighted by $\lambda_{\mathcal{F}}=0.1$ in Overcooked and $0.01$ in SMAX. Both native and generated shaping were applied with constant weights throughout training. The JaxMARL reference configuration for Overcooked instead decays its shaping during training, so the native baseline used here is the default reward under the same constant schedule as the generated reward.

\subsection{Policy Learning}
\label{subsec:learning}
Policies were trained with multi-agent proximal policy optimization (MAPPO) \cite{yu2022}, an actor--critic method for cooperative MARL, under centralized training and decentralized execution. Each agent selects its actions from its own observation through an actor network whose parameters are shared by all agents. A centralized critic, used only during training, estimates per-agent values from joint information and thereby supports credit assignment within the team.

\begin{figure*}[t]
\centering
\includegraphics[width=\textwidth]{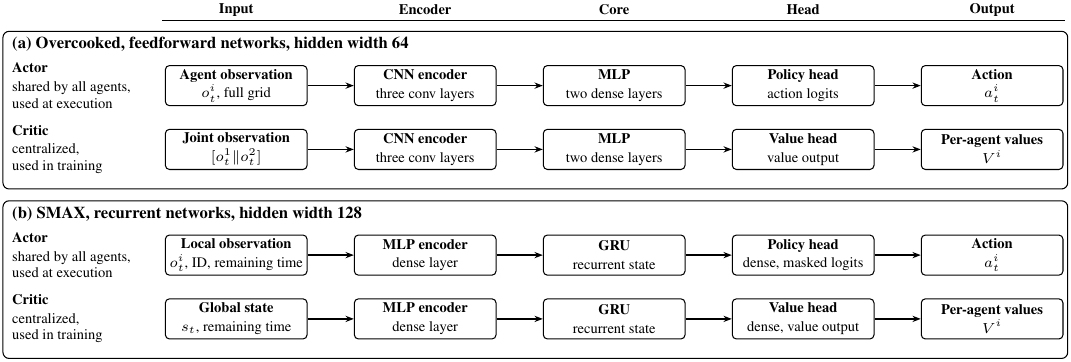}
\caption{MAPPO architectures. (a) Overcooked: feedforward convolutional networks on full-grid observations. (b) SMAX: recurrent networks with action masking. Actor parameters are shared across agents, and critics are used only during training.}
\label{fig:architectures}
\end{figure*}

The network architectures are shown in Fig.~\ref{fig:architectures}. In Overcooked, the full-grid observation of each agent is encoded by three convolutional layers followed by two dense layers, and the critic receives the concatenated observations of both agents. In SMAX, where each unit observes only its surroundings, recurrent networks with gated recurrent units (GRUs) are used to retain information over time. The actor receives the local observation, the agent identity, and the remaining episode time, and masks unavailable actions, whereas the critic receives the global state and the remaining time.

The same architecture, optimization settings, and training budget of approximately five million environment steps were used for every reward condition within a family, as listed in Table~\ref{tab:mappo}. Every policy was trained from random initialization, and the final policy was evaluated with sampled actions over 64 episodes in Overcooked and 256 episodes in SMAX. Held-out results were not used for checkpoint selection.

\begin{table*}[t]
\centering
\caption{MAPPO Training Settings}
\label{tab:mappo}
\footnotesize
\setlength{\tabcolsep}{8pt}
\renewcommand{\arraystretch}{1.1}
\begin{tabular}{@{}lcc@{\hspace{34pt}}lcc@{}}
\toprule
Setting & Overcooked & SMAX & Setting & Overcooked & SMAX\\
\midrule
Hidden width & 64 & 128 & PPO minibatches & 16 & 4\\
Environment steps per training & 4\,997\,120 & 4\,997\,120 & Initial learning rate & $5\times10^{-4}$ & $2\times10^{-3}$\\
Parallel environments & 64 & 128 & Entropy coefficient & 0.01 & 0\\
Rollout length & 256 & 128 & Gradient norm bound & 0.50 & 0.25\\
Episode limit & 400 & 100 & Evaluation episodes & 64 & 256\\
\bottomrule
\end{tabular}
\par\vspace{3pt}
\parbox{\textwidth}{\footnotesize Both families used 305 updates, 4 PPO epochs, discount 0.99, GAE parameter 0.95, clipping 0.20, value coefficient 0.50, and linear learning-rate decay. PPO, proximal policy optimization. GAE, generalized advantage estimation.}
\end{table*}

\subsection{Reward Search and Controls}
\label{subsec:conditions}
Each reward search consisted of $G=3$ generations with $P=2$ candidates per generation, so six candidate programs were developed and assessed for each family. Within each candidate session, at most two learning probes could be run, each being a complete training run with the fixed learner and budget. Probes and development assessments used a single training seed, and each session was limited by a fixed time and interaction allowance. The reward engineer was the model \texttt{gpt-6.1-sol} at maximum reasoning effort.

To examine the role of autonomous evidence access, two SMAX controls were evaluated with the same permitted initial source context, candidate and probe budgets, learner, and selection rule. Only the evidence available to the reward engineer and the use of reward history differed.

\textit{Fixed diagnostics.} Native-baseline diagnostics and completed earlier-generation designs were retained, but training evidence was restricted to standardized reports of returns, learning curves, episode lengths, losses, entropy, and shaping statistics.

\textit{Eureka-inspired reflection} \cite{ma2024}. The best reward of the previous generation was refined with standardized task and reward-component feedback, without initial native diagnostics.

In both controls, probe feedback was restricted in the same way, and shell access and custom analyses were disabled. Evidence access, history, and inference effort were therefore compared jointly rather than as isolated factors. One search was executed per condition, and all SMAX rewards were frozen before held-out training began.

\subsection{Evaluation Protocol}
\label{subsec:statistics}
Each reward was developed once, and every held-out condition was then trained with \policyseeds{} matched policy seeds, so that the same seeds were used for every reward. Performance is reported as the mean and standard error of the per-seed evaluation means, with standard error defined as the sample standard deviation divided by $\sqrt{\policyseeds}$. Overcooked returns are expressed as percentages of the observed native mean. Variation in policy training under one selected reward is captured by these seeds, whereas variation across reward searches is not.

Gains are reported as point estimates of the difference between the GARDEN mean and the reference mean on each held-out scenario. The gains over native shaping are given in Table~\ref{tab:transfer}, and those over the reward-design controls are given in Section~\ref{subsec:controls_results}.

The scenario split and training budget were fixed after pilot calibration, and one held-out SMAX scenario had been examined in pilots, so the study is a pilot-informed rather than a researcher-blind evaluation. All runs were executed on a single NVIDIA RTX 6000 Ada GPU, and the software configuration is provided in the repository.

\section{Experimental Results}
\label{sec:results}
\subsection{Transfer to Held-Out Scenarios}
\label{subsec:transfer_results}
The two families are reported on their natural scales. Overcooked task returns differ in magnitude between layouts, so they are expressed as percentages of the native result on each layout. Native shaping then corresponds to 100\%, and the gain of GARDEN is its relative improvement. SMAX outcomes are already win rates, so their gains are absolute differences in percentage points (pp). This reporting follows the selection score in \eqref{eq:selection_score}, which uses relative improvement for return-based tasks and absolute improvement for win rates.

\begin{table*}[t]
\centering
\caption{Held-Out Performance of Native Shaping and Frozen GARDEN Rewards}
\label{tab:transfer}
\small
\setlength{\tabcolsep}{9pt}
\renewcommand{\arraystretch}{1.15}
\begin{tabular}{@{}lcccc@{}}
\toprule
& \multicolumn{2}{c}{Overcooked, task return (\% of native)} & \multicolumn{2}{c}{SMAX, win rate (\%)}\\
\cmidrule(lr){2-3}\cmidrule(l){4-5}
Training reward & Forced Coordination & Counter Circuit & 10m vs 11m & SMACv2 10 units\\
\midrule
Native shaping & \ms{100.0}{3.4} & \ms{100.0}{4.3} & \ms{27.3}{0.96} & \ms{74.6}{1.80}\\
\textbf{GARDEN} & \bms{158.2}{6.8} & \bms{152.5}{15.8} & \bms{96.1}{0.90} & \bms{83.6}{0.86}\\
\midrule
Gain over native & $+58.2$\,\% & $+52.5$\,\% & $+68.8$\,pp & $+9.0$\,pp\\
\bottomrule
\end{tabular}
\par\vspace{3pt}
\parbox{\textwidth}{\footnotesize \spreadnote{} Overcooked means and standard errors are expressed as percentages of the observed native mean, which is treated as a fixed reporting scale, so Overcooked gains are relative improvements. SMAX gains are differences in win rate in percentage points (pp). Gains are reported as point estimates. Bold marks the best result.}
\end{table*}

\begin{table*}[t]
\centering
\caption{SMAX Held-Out Win Rate and LLM Usage of the Reward-Design Methods}
\label{tab:controls}
\small
\setlength{\tabcolsep}{9pt}
\renewcommand{\arraystretch}{1.15}
\begin{tabular}{@{}lccccc@{}}
\toprule
Method & 10m vs 11m & SMACv2 10 units & Mean & Model calls & Input tokens (M)\\
\midrule
Native shaping & \ms{27.3}{0.96} & \ms{74.6}{1.80} & 51.0 & -- & --\\
Fixed diagnostics & \ms{91.2}{0.74} & \ms{77.1}{0.88} & 84.2 & 18 & 3.5\\
Eureka-inspired reflection & \ms{87.6}{0.92} & \ms{70.8}{0.94} & 79.2 & 18 & 3.3\\
\textbf{GARDEN} & \bms{96.1}{0.90} & \bms{83.6}{0.86} & {\boldmath$89.9$} & 155 & 15.0\\
\bottomrule
\end{tabular}
\par\vspace{3pt}
\parbox{\textwidth}{\footnotesize Win rates in percent. \spreadnote{} The mean column is the equal-weight average of the two scenario means. Model calls are LLM requests during the search, and input tokens are given in millions. Bold marks the best result.}
\end{table*}

Native shaping and the frozen GARDEN rewards are compared in Table~\ref{tab:transfer}. Performance was improved by GARDEN in all four held-out scenarios. In Overcooked, task returns were increased by 58.2\% on Forced Coordination and by 52.5\% on Counter Circuit, with the same frozen program used on both layouts. In SMAX, win rates were increased by 68.8 pp against the larger marine team and by 9.0 pp in the randomized ten-unit battles, which raised the mean held-out win rate from 51.0\% to 89.9\%. The largest gain was obtained under numerical disadvantage, where native shaping was least effective. Because all held-out policies were trained from scratch with the frozen source, these results measure the transfer of the reward itself rather than of a trained policy.

\subsection{Comparison With Diagnostic and Reflection Controls}
\label{subsec:controls_results}
The SMAX comparison is reported in Table~\ref{tab:controls}. The highest win rate was achieved by GARDEN in both held-out scenarios and on average. On 10m vs 11m, all generated rewards exceeded native shaping by a wide margin. On the randomized SMACv2 10 units scenario, the Eureka-inspired reward fell below native shaping (70.8\% versus 74.6\%), whereas the largest improvement was obtained with GARDEN (83.6\%). Its mean held-out win rate of 89.9\% exceeded those of fixed diagnostics (84.2\%) and Eureka-inspired reflection (79.2\%) by 5.7 and 10.7 pp, respectively. Relative to fixed diagnostics, gains of 4.9 pp on 10m vs 11m and 6.5 pp on SMACv2 10 units were obtained. These comparisons hold the candidate and probe budgets, learner, and selection rule fixed. In the evaluated searches, the reward developed with autonomous access to source code, raw training artifacts, and custom analyses achieved higher held-out performance, with the additional model calls listed in Table~\ref{tab:controls}.

\subsection{Evidence-Guided Reward Development}
\label{subsec:evidence_cases}
Two development investigations illustrate how training evidence was used to construct and revise reward programs. These cases concern development drafts rather than the programs ultimately selected for transfer.

\textit{Overcooked: unused production capacity.} A custom trajectory analysis revealed that the second pot in Bottleneck remained empty throughout eight recorded 400-step native-policy episodes. Demand-aware shaping, wall-respecting routing, and partly shared production credits were subsequently incorporated into a reward program. Both pots were used in the recorded trajectories of this candidate, and the mean number of deliveries increased from nine to twelve. In the separate 64-episode evaluation, the mean task return increased by 33.5\%. These observations support increased use of the available capacity under the combined reward, without identifying which component produced the change.

\textit{SMAX: revision after an unsuccessful probe.} On 5m vs 6m, an initial reward draft achieved 22 wins in 256 evaluation episodes. Combat traces and missed firing opportunities were then examined. Individual survival penalties were removed, shared preservation was retained, and the offensive, evasion, and terminal terms were revised. Under the same training budget and seed, the revised draft achieved 238 wins in 256 episodes, which increased the win rate from 8.6\% to 93.0\%. The recorded investigation and revision preceded the improved result.

Each probe trained a new policy for 4\,997\,120 environment steps with one fixed training seed. These cases document how evidence was translated into reward revisions and tested through learning probes.

\section{Conclusion}
\label{sec:conclusion}
GARDEN was presented as an approach to autonomous reward engineering for cooperative task families. Rather than being designed separately for each scenario, one executable reward program was developed from several development scenarios. The investigation of environment source, training artifacts, and behavioral evidence was delegated to a tool-using LLM, and draft rewards were tested through learning probes. The learner and the selection rule were held fixed, and the selected source was frozen before it was used to train new policies on held-out scenarios.

Successful transfer of the frozen rewards was demonstrated in Overcooked and SMAX. Held-out task returns were increased by 58.2\% and 52.5\% in Overcooked, and the mean held-out win rate in SMAX was raised from 51.0\% to 89.9\%. Because the policies were trained from scratch on scenarios that were concealed during design, these gains reflect the generalization of the reward itself. Under matched candidate and probe budgets, GARDEN also outperformed fixed-diagnostic and Eureka-inspired reward design by 5.7 and 10.7 pp, supporting the effectiveness of the evaluated reward-design process.

The recorded development cases further showed how source inspection, custom analyses, and learning probes were used to diagnose coordination problems and to revise reward programs. Together, these results indicate that task-family reward engineering with autonomous evidence navigation is an effective route toward reusable reward design for cooperative multi-agent systems, and they motivate its extension to further task families and larger teams.

\section*{Code and Data Availability}
The implementation, prompts, and experimental configuration are available at \url{https://github.com/DoganUrgun/GARDEN}. The Overcooked and SMAX environments are part of the publicly available JaxMARL library~\cite{rutherford2024}. Other data supporting the findings are available from the author on reasonable request.

\IEEEtriggeratref{17}


\begin{thebibliography}{99}

\bibitem{gronauer2022}
S.~Gronauer and K.~Diepold, ``Multi-agent deep reinforcement learning: A survey,'' \emph{Artif. Intell. Rev.}, vol.~55, no.~2, pp.~895--943, 2022, doi: 10.1007/s10462-021-09996-w.

\bibitem{zhu2025}
G.~Zhu, R.~Zhou, W.~Ji, and S.~Zhao, ``LAMARL: LLM-aided multi-agent reinforcement learning for cooperative policy generation,'' \emph{IEEE Robot. Autom. Lett.}, vol.~10, no.~7, pp.~7476--7483, Jul. 2025, doi: 10.1109/LRA.2025.3577527.

\bibitem{wei2026pcrd}
D.~Wei, P.~Yi, J.~Lei, Y.~Hong, H.~Dong, and Y.~Du, ``An automated reinforcement learning reward design framework with large language model for cooperative platoon coordination,'' \emph{IEEE Trans. Intell. Transp. Syst.}, vol.~27, no.~7, pp.~8252--8269, Jul. 2026, doi: 10.1109/TITS.2026.3664858.

\bibitem{wong2023}
A.~Wong, T.~B\"{a}ck, A.~V. Kononova, and A.~Plaat, ``Deep multiagent reinforcement learning: Challenges and directions,'' \emph{Artif. Intell. Rev.}, vol.~56, pp.~5023--5056, 2023, doi: 10.1007/s10462-022-10299-x.

\bibitem{ng1999}
A.~Y. Ng, D.~Harada, and S.~Russell, ``Policy invariance under reward transformations: Theory and application to reward shaping,'' in \emph{Proc. 16th Int. Conf. Mach. Learn.}, 1999, pp.~278--287.

\bibitem{hadfield2017}
D.~Hadfield-Menell, S.~Milli, P.~Abbeel, S.~Russell, and A.~D. Dragan, ``Inverse reward design,'' in \emph{Proc. Adv. Neural Inf. Process. Syst.}, vol.~30, 2017.

\bibitem{xie2024}
T.~Xie \emph{et al.}, ``Text2Reward: Reward shaping with language models for reinforcement learning,'' in \emph{Proc. Int. Conf. Learn. Represent.}, 2024.

\bibitem{ma2024}
Y.~J. Ma \emph{et al.}, ``Eureka: Human-level reward design via coding large language models,'' in \emph{Proc. Int. Conf. Learn. Represent.}, 2024.

\bibitem{li2026arbo}
M.~Li, R.~Tan, and R.~Wang, ``Beyond scripted search: Sample-efficient reward discovery via agentic black-box optimization,'' 2026, \emph{arXiv:2609.32394}, doi: 10.48550/arXiv.2609.32394.

\bibitem{zeng2024}
Y.~Zeng, Y.~Mu, and L.~Shao, ``Learning reward for robot skills using large language models via self-alignment,'' in \emph{Proc. 41st Int. Conf. Mach. Learn.}, ser. Proc. Mach. Learn. Res., vol.~235, 2024, pp.~58366--58386.

\bibitem{gao2025}
N.~Gao, X.~Zhang, X.~Jiang, M.~You, M.~Zhang, and Y.~Deng, ``RF-Agent: Automated reward function design via language agent tree search,'' in \emph{Proc. Adv. Neural Inf. Process. Syst.}, vol.~38, 2025.

\bibitem{lee2026}
H.~Lee \emph{et al.}, ``Reward design agent for reinforcement learning,'' \emph{Reinforcement Learn. J.}, vol.~7, 2026. [Online]. Available: \url{https://rlj.cs.umass.edu/2026/papers/Paper81.html}

\bibitem{abbeel2004}
P.~Abbeel and A.~Y. Ng, ``Apprenticeship learning via inverse reinforcement learning,'' in \emph{Proc. 21st Int. Conf. Mach. Learn.}, 2004, doi: 10.1145/1015330.1015430.

\bibitem{christiano2017}
P.~F. Christiano, J.~Leike, T.~B. Brown, M.~Martic, S.~Legg, and D.~Amodei, ``Deep reinforcement learning from human preferences,'' in \emph{Proc. Adv. Neural Inf. Process. Syst.}, vol.~30, 2017.

\bibitem{zou2021}
H.~Zou, T.~Ren, D.~Yan, H.~Su, and J.~Zhu, ``Learning task-distribution reward shaping with meta-learning,'' in \emph{Proc. AAAI Conf. Artif. Intell.}, vol.~35, no.~12, 2021, pp.~11210--11218, doi: 10.1609/aaai.v35i12.17337.

\bibitem{foerster2018}
J.~Foerster, G.~Farquhar, T.~Afouras, N.~Nardelli, and S.~Whiteson, ``Counterfactual multi-agent policy gradients,'' in \emph{Proc. AAAI Conf. Artif. Intell.}, vol.~32, no.~1, 2018, doi: 10.1609/aaai.v32i1.11794.

\bibitem{rashid2018}
T.~Rashid, M.~Samvelyan, C.~Schroeder, G.~Farquhar, J.~Foerster, and S.~Whiteson, ``QMIX: Monotonic value function factorisation for deep multi-agent reinforcement learning,'' in \emph{Proc. 35th Int. Conf. Mach. Learn.}, ser. Proc. Mach. Learn. Res., vol.~80, 2018, pp.~4295--4304.

\bibitem{li2025rstar}
P.~Li \emph{et al.}, ``R*: Efficient reward design via reward structure evolution and parameter alignment optimization with large language models,'' in \emph{Proc. 42nd Int. Conf. Mach. Learn.}, ser. Proc. Mach. Learn. Res., vol.~267, 2025, pp.~34509--34527.

\bibitem{qu2025}
Y.~Qu \emph{et al.}, ``Latent reward: LLM-empowered credit assignment in episodic reinforcement learning,'' in \emph{Proc. AAAI Conf. Artif. Intell.}, vol.~39, no.~19, 2025, pp.~20095--20103, doi: 10.1609/aaai.v39i19.34213.

\bibitem{bai2026}
J.~Bai \emph{et al.}, ``A hierarchical multi-agent reinforcement learning framework with high-level guidance from large language models,'' \emph{Sci. Rep.}, vol.~16, Art.~no.~23832, 2026, doi: 10.1038/s41598-026-54971-6.

\bibitem{urgun2026}
D.~Urgun and G.~Gungor, ``Large language model guided incentive aware reward design for cooperative multi-agent reinforcement learning,'' 2026, \emph{arXiv:2603.24324}, doi: 10.48550/arXiv.2603.24324.

\bibitem{wu2026}
X.~Wu, Z.~Zheng, and H.~Xiong, ``LLM-ALSO: LLM-driven adaptive learning-signal optimization for multi-agent reinforcement learning,'' 2026, \emph{arXiv:2605.29293}, doi: 10.48550/arXiv.2605.29293.

\bibitem{ellis2023}
B.~Ellis \emph{et al.}, ``SMACv2: An improved benchmark for cooperative multi-agent reinforcement learning,'' in \emph{Proc. Adv. Neural Inf. Process. Syst.}, vol.~36, 2023, pp.~37567--37593.

\bibitem{ruhdorfer2025}
C.~Ruhdorfer, M.~Bortoletto, A.~Penzkofer, and A.~Bulling, ``The Overcooked generalisation challenge: Evaluating cooperation with novel partners in unknown environments using unsupervised environment design,'' \emph{Trans. Mach. Learn. Res.}, 2025.

\bibitem{oliehoek2016}
F.~A. Oliehoek and C.~Amato, \emph{A Concise Introduction to Decentralized POMDPs} (SpringerBriefs in Intelligent Systems). Cham, Switzerland: Springer, 2016, doi: 10.1007/978-3-319-28929-8.

\bibitem{rutherford2024}
A.~Rutherford \emph{et al.}, ``JaxMARL: Multi-agent RL environments and algorithms in JAX,'' in \emph{Proc. Adv. Neural Inf. Process. Syst.}, vol.~37, 2024.

\bibitem{carroll2019}
M.~Carroll \emph{et al.}, ``On the utility of learning about humans for human-AI coordination,'' in \emph{Proc. Adv. Neural Inf. Process. Syst.}, vol.~32, 2019, pp.~5174--5185.

\bibitem{samvelyan2019}
M.~Samvelyan \emph{et al.}, ``The StarCraft multi-agent challenge,'' in \emph{Proc. 18th Int. Conf. Auton. Agents MultiAgent Syst.}, 2019, pp.~2186--2188.

\bibitem{yu2022}
C.~Yu \emph{et al.}, ``The surprising effectiveness of PPO in cooperative multi-agent games,'' in \emph{Proc. Adv. Neural Inf. Process. Syst.}, vol.~35, 2022, pp.~24611--24624.

\end{thebibliography}
\end{document}